# GreekT5: A Series of Greek Sequence-to-Sequence Models for News Summarization

Nikolaos Giarelis*

University of Patras, giarelis@ceid.upatras.gr

Charalampos Mastrokostas

University of Patras, cmastrokostas@ac.upatras.gr

Nikos Karacapilidis

University of Patras, karacap@upatras.gr

Text summarization (TS) is a natural language processing (NLP) subtask pertaining to the automatic formulation of a concise and coherent summary that covers the major concepts and topics from one or multiple documents. Recent advancements in deep learning have led to the development of abstractive summarization transformer-based models, which outperform classical approaches. In any case, research in this field focuses on high resource languages such as English, while the corresponding work for low resource languages is still underdeveloped. Taking the above into account, this paper proposes a series of novel TS models for Greek news articles. The proposed models were thoroughly evaluated on the same dataset against GreekBART, which is the state-of-the-art model in Greek abstractive news summarization. Our evaluation results reveal that most of the proposed models significantly outperform GreekBART on various evaluation metrics. We make our evaluation code public, aiming to increase the reproducibility of this work and facilitate future research in the field.

CCS CONCEPTS • Natural Language Processing • Machine Learning • Deep Learning

**Additional Keywords and Phrases:** Greek Language, Text Summarization, Pretrained Language Models

## 1 INTRODUCTION

Natural language processing (NLP) is an interdisciplinary research field, which incorporates aspects and approaches from the fields of computer science, artificial intelligence and linguistics; it is concerned with the development of approaches that efficiently and semantically process large volumes of textual data. Text summarization (TS) is a fundamental NLP subtask, which has been defined as the process of automatically creating a succinct and coherent summary that captures the principal ideas and topics from one or multiple documents (El-kassas et al., 2021). The task of manually summarizing large textual resources by human authors is time-consuming and monotonous (Gupta and Lehal, 2010), which calls for the automation of the summarization task. As mentioned in a recent TS survey (Giarelis et al., 2023), diverse TS applications for the English language already exist, including: (i) short document TS (e-mail, news); (ii) long document TS (books, legal texts), and (iii) TS from extremely short texts (argument summarization for measuring public opinion in digital democracy platforms; Adamides et al., 2023).

* Corresponding Author.

Several TS reviews (El-kassas et al., 2021; Alomari et al. 2022; Giarelis et al. 2023) mention that the TS approaches can be classified into two major types, namely extractive and abstractive, the former focus on extracting important sentences from the text, while the latter generate the text in an abstractive manner similarly to the paraphrasing process of a human author; this leads to the abstractive TS approaches producing more coherent summaries than their extractive counterparts. In recent TS reviews (Alomari et al., 2022; Giarelis et al., 2023), abstractive TS approaches that rely on deep learning (DL) techniques have been shown to achieve state-of-the art results, outperforming the majority of extractive ones. Specifically, Alomari et al. (2022) evaluate many abstractive TS approaches that are based on various DL architectures (e.g., RNN, LSTM, GRU, etc.), showcasing that the approaches built on the Transformer-based sequence-to-sequence (*Seq2Seq*) architecture (Song et al., 2019) achieve state-of-the-art performance. The transformer-based *Seq2Seq* models utilize the encoder-decoder architecture, where an encoder encodes a sequence of textual tokens into a semantic representation, then a decoder decodes this representation into a sequence of tokens. Both the encoder and decoder are trained as to maximize the probability of reconstructing the given output sequences based on the given input sequences. In this work, we focus on abstractive TS approaches, built on *Seq2Seq*.

In any case, research in this field focuses on high resource languages such as English, while the corresponding work for low resource languages is still underdeveloped; in our case, the corresponding Greek TS literature which is rather limited. This is due to the fact that state-of-the-art DL techniques for the Greek Language are underdeveloped (Evdaimon et al., 2023), while for model training only a few Greek NLP datasets and resources exist (Papantoniou and Tzitzikas, 2020). Earlier introduced DL models for the Greek language, namely *GreekBert* (Koutsikakis et al., 2020) and *GreekLegalBert* (Athinaios et al., 2020) are based on the *BERT* architecture (Devlin et al., 2019), which lacks text generative capabilities. For this reason, Greek *BERT*-based models are not suited for abstractive TS. On the contrary, the *Seq2Seq* architecture facilitates the development of DL models that are able to handle text generation tasks (e.g., summarization, translation etc.).

Considering the above remarks, this paper aims to advance the state-of-the-art in Greek NLP models, putting emphasis on DL methodologies for TS. Specifically, we propose a series of novel TS models for the summarization of Greek news articles. The proposed models were thoroughly evaluated on the same news summarization dataset, namely *GreekSUM*, against *GreekBART*, which is the first Greek sequence-to-sequence model and state-of-the-art in Greek abstractive news summarization (Evdaimon et al., 2023). Our evaluation results reveal that most of our models significantly outperform *GreekBART* on various evaluation metrics. Aiming to increase the reproducibility of our work and facilitate future research in the field, we make our evaluation code (https://github.com/NC0DER/GreekT5) and models (https://huggingface.co/IMISLab) public.

The remainder of this paper is organized as follows. A list of recent language model architectures that can be adapted for the Greek language is presented in Section 2. In the same section, TS approaches that are built on these architectures for specific languages and domains are also presented. The proposed series of models, called *GreekT5*, along with the experimental setup and the evaluation process are thoroughly described in Section 3. Concluding remarks, offering a set of insights on current NLP methodologies and models for the Greek language, as well as future research directions are outlined in Section 4.

## 2 RELATED WORK

In this section, we present a set of DL architectures (language models) that are pre-trained on multilingual corpora. These language models enable researchers to finetune them for various downstream NLP tasks (e.g., TS), using datasets from different languages and of diverse topics. Section 2.1 reports on prominent multilingual language model architectures that are used as a basis for finetuning, while Section 2.2 presents a series of related TS abstractive models for specific languages and domains.

### 2.1 Multilingual Language Models

The language models presented below facilitate the finetuning process. They are pre-trained on large multilingual corpora and are used as a training base for finetuning. It is imperative that these models are finetuned for a specific combination of languages and downstream tasks on appropriate datasets, otherwise they cannot function.

*mBART* (Liu et al., 2020) is a multilingual pretrained *Seq2Seq* language model with a denoising auto-encoder architecture. This model has the same pretraining scheme as *BART* (Lewis et al., 2019), applied to large-scale multilingual corpora, since *BART* was pretrained only for English. Specifically, *mBART* has 12 layers each for its encoder and decoder, with a model dimension of 1024, 16 attention heads and an estimated size of 680 million parameters. The model also includes an extra normalization layer, which was introduced to stabilize the training process when using 16-bit floating precision (*FP16*).

Other *Seq2Seq* models apply the masked language modelling technique (*MLM*), so their decoder learns to predict tokens masked by the encoder, in contrast *mBART* reconstructs the full target sequence, which allows the application of *MLM*, as well as selected noise functions. In particular, *mBART* uses two noise functions; the first one is masking phrases, while the second is permuting sentences of the input dataset. Then the *Seq2Seq* hidden layer learns to recreate text sequences by "denoising" them. *mBART* was pretrained for multiple languages using the *CC25* (Wenzek et al., 2019) dataset, which contains documents from 25 different languages from Common Crawl. The authors have also published an updated *mBART* model (https://huggingface.co/facebook/mbart-large-50), which supports 50 natural languages. Unfortunately, none of these two models can be finetuned on Greek NLP tasks, since their pretraining datasets do not include Greek.

*mt5* (Xue et al., 2021) is a multilingual pretrained *Seq2Seq* language model, based on the pretraining scheme as *T5* (Raffel et al., 2020). Both models follow the "text-to-text" strategy, in which the same training hyperparameters are used by the model for similar text generative tasks (e.g., TS, translation etc.) at a zero-shot fashion using task prefixes (e.g., *"summarize: <input_text>"* or *"translate from English to Greek: <input_text>")*. Moreover, *mT5* possesses a customized *MLM* learning object called "span-corruption", where instead of masking input tokens, the model masks spans of input tokens and learns to predict them.

For the pretraining of *mt5*, a multilingual corpus - *mC4* (Xue et al., 2021) - comprising of 101 natural languages collected from Common Crawl is utilized. The authors also utilize a language sampling strategy, since not all languages are equally represented in the dataset. This strategy involves collecting more training examples for low resource languages, to avoid underfitting for these languages. The *mC4* dataset contains 43 billion Greek tokens, thus *mt5* supports finetuning for Greek NLP. *mt5* is distributed in five variants, namely *mt5-small*, *mt5-base*, *mt5-large*, *mt5-xl* and *mt5-xxl*. The authors report that even the small variant has many more parameters than previous language models. They also evaluate these variants on the question answering NLP task and their results indicate that beyond the base model, there is a tradeoff of diminishing returns in accuracy gain (up to *+4%* F1 score increase) while the parameter number of variants increases (e.g.,

*mt5-base*: 580 million parameters vs. *mt5-xxl*: 13 billion parameters). This significant increase in model size leads to exceedingly more computational resources required to run the larger variants.

*umt5* (Chung et al., 2023) is an improvement on *mt5*, which updates the language sampling strategy into a more efficient one (*UNIMAX*) for low resource languages. The *UNIMAX* (uniform + max) strategy assumes that large-scale training jobs operate with a fixed computational amount. Initially, *UNIMAX* starts by pre-allocating training tokens to underrepresented linguistic datasets based on a set number of allowed max repeats (*N*) for tokens of each language. The remaining token budget is allocated uniformly across all sufficiently represented languages to avoid surpassing the predetermined number of epochs for each language. Unlike previous sampling strategies, *UNIMAX* is relatively resistant to distribution biases of a multilingual corpus. Case in point, the learning strategy of *mt5* samples more tokens for an overrepresented language (e.g., English) than underrepresented languages, while *UNIMAX* assigns equal training tokens for each language if it does not repeat tokens more than N times. The authors of *umt5* argue that since the proposed strategy directly controls the token amount sampled for each language, it prevents overfitting on low-resource languages, without imposing any reprioritization on higher-resource languages. The authors introduce an updated version of *mC4* which comprises 29 trillion characters across 107 languages, while releasing a new series of pretrained language models for finetuning (i.e., *umt5-small, umt5-base, umt5-xl, umt5-xxl*) that outperform their *mt5* counterparts.

### 2.2 Abstractive Text Summarization Models

Using the language models presented in Section 2.1 as a basis, this section reports on abstractive models pretrained and / or finetuned for TS for various natural languages, alongside their employed datasets.

*WATS-SMS* (Fendji et al., 2021) is a French *Seq2Seq* model finetuned on *T5*, for summarizing webpages into an SMS format. This summary is meant for users accessing webpages on areas with limited mobile connectivity. The model was finetuned on a dataset consisting of 15.000 French Wikipedia articles, where the summary section of each article was used as the reference summary.

*NASca* and *NASes* (Ahuir et al., 2021) are two pretrained and finetuned *Seq2Seq* models, based on *BART* for Spanish and Catalan news summarization. For the pretraining on *NASca*, a textual corpus was used, which comprised 9.3GB of Catalan documents from the news dataset, the *OSCAR* (Juarez et al., 2020) corpus and the Catalan version of Wikipedia, while for *NASes*, Spanish documents from the news dataset were used alongside Spanish Wikipedia articles. *mBART* and *mt5* were also finetuned on the previous datasets for Spanish and Catalan, respectively.

Eddine et al., (2022) train and finetune a series of *Seq2Seq* abstractive TS models for Arabic. The first introduced model is *AraBART*, which uses the same pretraining scheme as *BART*. *AraBART* is pretrained on a 20GB corpus of Arabic documents with a vocabulary size of 50 thousand tokens. The authors of *AraBART* also finetuned the base variants of *mBART* and *mt5*, and *AraT5* (Al-Maleh and Desouki, 2020) on various news summarization datasets. The authors reported that *AraBART* achieves better performance than the other considered Arabic models.

Quatra and Cagliero (2022) train and finetune a series of *Seq2Seq* abstractive TS models for the Italian language. The first introduced model is *BART-IT*, which uses the same pretraining scheme as *BART*. *BART-IT* is pretrained on a large textual corpus from the Italian version of *mC4* containing 103 million documents and 41 billion words after preprocessing is applied. This model has a vocabulary size of 52.000 Italian tokens. The authors also finetune *BART-IT*, *mT5*, and *mBART* and compare their performance across multiple news summarization datasets, and a dataset containing abstracts from the Italian Wikipedia.

To the best of our knowledge, there are no abstractive summarization approaches for the Greek language based on *Seq2Seq*, other than *GreekBART* (Evdaimon et al., 2023), which is the first pretrained Greek *Seq2Seq* model based on *BART*. This model is pretrained on an 87.6 GB Greek corpus, with a vocabulary of 50,000 tokens extracted from a 20 GB random corpus sample. This corpus was made using multiple sources including the Greek Part of Wikipedia, the Greek Part of the *OSCAR* corpus (Abadji et al., 2022), the Greek part of the European Parliament Proceedings Parallel Corpus (Koehn et al., 2005) and the Greek web corpus dataset (Outsios et al., 2018), The authors report that the corpus includes text with different Greek (formal and informal) writing styles (encyclopedic, political, journalistic etc.), as to make the pretraining of *GreekBART* more robust. The authors also pre-processed the pretraining corpus by removing URLs, emojis, tags, and hashtags, and sentences with no additional contextual meaning. They also removed short documents and duplicated information.

For the pretraining of *GreekBART*, the authors utilized the *BART* base architecture, which has an encoder and a decoder with 6 layers respectively, a hidden state with 768 layers, 12 attention heads in both the encoder and the decoder, and a normalization layer on both the encoder and the decoder. This resulted in a model of approximately 181 million parameters. The authors also used *FP16* to limit the memory and time requirements of the training process. For their pretraining hyperparameters the authors used an Adam optimizer with $e = 1e{-}6$, $\beta_1 = 9e{-}1$, and $\beta_2 = 999e{-}3$, with a learning rate 6e−4 with a linear decay, with a warm-up period of 6% of the overall training, while for the first 12 epochs, they fixed the dropout to 1e−1, whereas for epochs 12 to 16 it was reduced 5e−2, and finally was set to zero for epochs 16 to 20. The authors also pretrained and finetuned *GreekBART*, *mBART25* and *mBART50* on *GreekSUM* (Evdaimon et al., 2023); a Greek news summarization dataset discussed in Section 3.2. They used the same set of hyperparameters for finetuning all models; specifically, 30 training epochs with a learning rate of 5e−5 with a linear decay and a warmup period of 6% of the overall training. *GreekBART* achieved state-of-the-art results for Greek TS.

## 3 GREEKT5: A SERIES OF ABSTRACTIVE MODELS FOR GREEK NEWS SUMMARIZATION

This section provides details regarding our approach for Greek abstractive TS, as well as the experimental setup. Specifically, Section 3.1 reports on the hardware and software specifications employed in our research work; Section 3.2 describes the selected dataset and metrics for our finetuning and evaluation processes; Section 3.3 describes our approach for finetuning language models for the Greek language; Section 3.4 presents evaluation results concerning *GreekT5*, *GreekBART* and an extractive baseline model.

### 3.1 Hardware and Software Specifications

The finetuning process was conducted on a computer using an Intel Core i9-12900K 16-Core CPU with 64 GB of RAM and an Nvidia RTX A400 16 GB VRAM GPU, while the evaluation process was conducted on a computer using an AMD Ryzen 5 6-Core CPU with 16 GB of RAM and an Nvidia RTX 3060 12 GB VRAM GPU. The use of GPUs is recommended for the acceleration of both processes, benefiting from runtime improvements offered by popular ML libraries (e.g., PyTorch, transformers).

Regarding the software specifications, we opted for Hugging Face's transformer library (https://huggingface.co/docs/transformers/) for the model finetuning process. For the experimental setup we utilized *TextRank* (Mihalcea and Tarau, 2004), a notable extractive TS approach as a baseline, and specifically its implementation (https://pypi.org/project/sumy/) from *sumy* (Belica, 2021). For the experimental setup, we also employ the *GreekBART* abstractive TS models for abstract summarization, which can be accessed upon request (http://nlp.polytechnique.fr/resources-greek#greek). Finally, for our evaluation we use these evaluation metric implementations for *BERTScore* (https://huggingface.co/spaces/evaluate-metric/bertscore) and *ROUGE* (https://huggingface.co/spaces/evaluate-metric/rouge).

### 3.2 Dataset and Evaluation Metrics

To the best of our knowledge, the only abstractive TS dataset for the Greek Language is *GreekSUM* (https://github.com/iakovosevdaimon/GreekSUM/). This dataset contains ~151,000 news articles collected from *News24/7* (https://www.news247.gr/), belonging to various topics (i.e., society, politics, economy, culture or world news). Every article is accompanied by a one-sentence title and an abstractive summary, with an average length of 9.95 and 24.55 words respectively. Overall, the dataset has a train-test-validation split (~87%, ~6.5%, ~6.5%). These two summary types lead the authors of *GreekSUM* to name these the two dataset versions (title vs abstract summarization) as *GreekSUM Title* and *GreekSUM Abstract*. For our experiments, we use the *GreekSUM Abstract* version, to finetune our models for the abstract TS task. For this task, we use its train split for finetuning, while for the evaluation of our models we are using its test split.

The automatic evaluation of TS models relies on comparing the textual similarity between machine generated summaries and human reference summaries, also referred to as "gold summaries". The most common evaluation metric of TS literature is *ROUGE* (Lin, 2004), which calculates various scores that measure the *n*-gram overlap between the generated and reference summaries. Similarly to other works presented in Section 2.2, we utilize the *ROUGE-1*, *ROUGE-2* and *ROUGE-L* scores, which respectively measure the number of common unigrams, bigrams and the longest common subsequence between the generated and reference summaries, thus computing an exact match similarity score. As mentioned in a recent TS review (Giarelis et al., 2023), exact match metrics such as *ROUGE* penalize highly abstractive summaries with synonymous terms, since they do not capture semantic similarity between the phrases of machine generated and gold summaries. To overcome this limitation, we also utilize *BERTScore* (Zhang et al., 2020), a state-of-the-art TS metric that measures this semantic similarity by calculating the cosine similarity between the token embeddings for each pair of evaluated summaries. In a non-English evaluation setting, the authors of *BERTScore* recommend the use of *mBERT* (Devlin et al., 2020), which is a pretrained multilingual encoder model, capable of generating token embeddings.

### 3.3 Finetuning

In the proposed *GreekT5* approach, we finetune, for the Greek language, three multilingual *Seq2Seq* models from *Hugging Face* (https://huggingface.co/), namely *google/mt5-small*, *google/umt5-small* and *google/umt5-base*. The selected models are well suited for this purpose, since their pretraining includes the Greek language, thus facilitating the finetuning for the downstream task of Greek TS. These models are distinguished by their model approach (e.g., *mt5* vs *umt5*) and their size indicator (e.g., *small* vs *base*), which refers to their relative size in terms of training parameters. Specifically, the size for the small models is 300 million parameters (~1.20 GB), while the base model has 580 million parameters (~2.37 GB). The proposed models produced after the finetuning process are referred to as *GreekT5* (*mt5-small*), *GreekT5* (*umt5-small*) and *GreekT5* (*umt5-base*).

For our finetuning process, we used *Hugging Face's* training framework for the models on the *GreekSUM* train split. For evaluation fairness, the proposed models were finetuned with the same training hyperparameters described below. The GPU batch size for training was set to 6 for both small models, while for the *google/umt5-base* model the parameter was set to 1 due to VRAM limitations. For finetuning, the number of total epochs was set to 10, while an AdamW optimizer was used with $e$ = 1e−8, $\beta_1$ and $\beta_2$ set to 0.9 and 0.0999 respectively, an initial learning rate of 3e−4 with linear weight decay and no warmup steps. All models were finetuned with 32-bit floating precision. Regarding the dataset tokenization, the maximum sequence length was set to 1024 tokens, due to the training limitations imposed by the *T5*-based models, while the maximum output length was set to 128 tokens. The tokenizer padding was set to *'max_length'* and truncation was enabled to handle longer sequences, which could not be processed within the limited input sequence length. Since *T5*-based models follow a multi-task architecture, the prefix *'summarize'* was prepended at the start of each training example, to denote to each model that it is finetuned for the TS task. No further data pre-processing procedures were carried out prior to finetuning.

### 3.4 Evaluation

For evaluation purposes, we utilized a set of extractive baseline models that are used by other summarization works, such as the ones presented in Section 2. These include *TextRank* (Mihalcea and Tarau, 2004), a notable extractive TS approach, and *LEAD-N* (Narayan et al., 2018), which extracts the first *N* sentences from the text.

To evaluate the proposed *GreekT5* models against *GreekBART* and the extractive baselines, the test split of the *GreekSUM Abstract* was used. This split consists of 10,000 pairs of news articles (full texts) and their respective human-assigned summaries (abstracts). The machine generated summaries of each approach were evaluated against the ones created by human authors. Considering that the abstract summaries of *GreekSUM* have an average number of 1.46 sentences, we set all extractive baselines to extract the top ranked sentence from each article. Regarding evaluation metrics, we opt for *ROUGE-1*, *ROUGE-2*, *ROUGE-L* and *BERTScore*. For each metric, we calculate the macro (mean) *F1* score of the scores computed for each summary pair in the test split. Tables 1 summarizes our experimental results; the best abstractive model scores for each metric are highlighted in bold.

Table 1: Experimental Evaluation Results on *GreekSUM Abstract* (macro F1 %).

| Approach | ROUGE-1 | ROUGE-2 | ROUGE-L | BERTScore |
|---|---|---|---|---|
| LEAD-1 | 25.51 | 11.33 | 20.16 | 72.66 |
| TextRank | 18.10 | 5.76 | 13.84 | 68.39 |
| GreekT5 (*mt5-small*) | 14.84 | 1.68 | 12.39 | 72.96 |
| GreekT5 (*umt5-small*) | 25.49 | 12.03 | 21.32 | 72.86 |
| GreekT5 (*umt5-base*) | **26.67** | **13.00** | **22.42** | 73.41 |
| GreekBART | 17.43 | 2.44 | 15.08 | **75.89** |

As shown in Table 1:

- *GreekT5 (umt5-base)* outperformed all models on most evaluation metrics except *BERTScore*, where *GreekBART* performed slightly better.
- The results for *GreekT5 (umt5-small)* are close to the results of the *GreekT5 (umt5-base)* model, showcasing small accuracy gains, despite the models' difference in terms of size.
- *GreekT5 (mt5-small)* achieves the lowest *ROUGE* scores; however, its *BERTScore* results are similar to those obtained from other abstractive models.
- *LEAD-1* achieves high scores, which indicates that a lot of summaries in the dataset are highly similar to the first sentence of the associated documents.
- *TextRank* remains a good extractive baseline model achieving better *ROUGE* scores than *GreekBART* and *GreekT5 (umt5-small)*; however, it achieves the lowest *BERTScore.*
- The best of the proposed models is *GreekT5 (umt5-base)*, which achieves the highest *ROUGE* scores, demonstrating its extractive potential (it is noted that it strongly outperforms *GreekBART*). It also achieves the second higher *BERTScore* (that is close to *GreekBART's* score), which demonstrates its abstractive capabilities.

A manual observation of the summaries produced by the models (see examples in *Appendix A*), validates the results shown in Table 1.

## 4 DISCUSSION

This paper has introduced a series of novel Greek abstractive TS models, based on multilingual models of the T5 architecture. As shown above, models based on *umT5* perform far better than *GreekBART* on the family of ROUGE metrics, while also demonstrate very good performance as far as the *BERTScore* metric is concerned. Concluding remarks are as follows:

- Since the introduction of the transformer-based *Seq2Seq* architecture, there is a paradigm shift towards DL for abstractive TS.
- Small model variants can perform well when finetuned for downstream tasks (e.g., TS), while having better performance than larger models.
- As shown in the related works presented in Section 2 and our experimental results presented in Section 3.4, we argue that multilingual *Seq2Seq* models, finetuned for a specific language and task, can achieve similar or even better performance compared to monolingual models pretrained and finetuned for the same language and task, while requiring significantly less computational resources.
- As shown in its original publication and our experimental results, the uniform sampling strategy of *umT5* for low resource languages does result in significant accuracy benefits compared to older language models finetuned using the same training data or hyperparameters.
- Our approach could be used for finetuning Greek *Seq2Seq* models for other domains (e.g., legal, medical, financial etc.) and tasks.

Our approach has the following limitations:

- Our series of models are finetuned in a news summarization dataset, which means that they might not be suitable for other domains or language settings that do not resemble the language style of the training corpus.
- *Seq2Seq* abstractive TS models require sufficient training data from different domains, which comes in contrast with the lack of human labelled datasets for text summarization, even more so in the case of the Greek language.
- Current abstractive *Seq2Seq* TS models exhibit certain limitations regarding their input length. Specifically, most of these models support a context window of 1024 input tokens, which hinders long document summarization (e.g., book, academic paper, financial reports summarization).
- These models also exhibit certain limitations regarding their output length, notably when increasing the model output (e.g., from 128 – 256 tokens). In this case, memory and training time requirements are increased substantially. Given the long training period of these models, which requires one or multiple GPUs, while factoring in their limited memory (VRAM) and increased costs, this hinders the overall development of certain DL models for certain application areas. Thus, making more efficient model architectures while democratizing GPU hardware is much needed for DL research.

Finally, open issues and future research directions include:

- The creation of more Greek datasets for abstractive TS, including texts belonging to diverse language styles and terminology (e.g., legal, medical, financial etc.), as to finetune more domain-specific TS models.
- The pre-training and/or fine-tuning of newer DL language models for the Greek language on multiple domains and NLP tasks.

- Human evaluation of DL models aiming to reveal cases where *Seq2Seq* TS models fail to generate proper summaries that are close to the gold ones; such an evaluation will lead to the improvement of existing TS models.
- Work towards the development of a general purpose TS model for the Greek language, which may be used in various settings.

## ACKNOWLEDGMENTS

We would like to thank Prof. Michalis Vazirgiannis for providing us access to the resources of *GreekBART* (hosted at http://nlp.polytechnique.fr/resources-greek).

# A APPENDIX

## Example 1

| Article | |
|---|---|
| Σε δημόσια διαβούλευση τίθεται εντός της ημέρας το νέο πλαίσιο για την συνολική διευθέτηση οφειλών και την παροχή δεύτερης ευκαιρίας για νοικοκυριά και επιχειρήσεις από το υπουργείο οικονομικών. Το νέο πλαίσιο φιλοδοξεί να αποτελέσει μια ολιστική λύση για τη συνολική διευθέτηση του ιδιωτικού χρέους με γρήγορες και απλές διαδικασίες από τους ενδιαφερόμενους. Σε δήλωσή του με την ευκαιρία της δημοσιοποίησης του σχεδίου νόμου το οποίο θα αναρτηθεί και θα είναι διαθέσιμο για διαβούλευση σε δύο εβδομάδες ο *υπουργός* οικονομικών κ. Χρήστος Σταϊκούρας περιγράφει τις 12 βασικές αλλαγές που φέρνει το νέο πλαίσιο. Συγκεκριμένα ο υπουργός οικονομικών αναφέρει ότι με τον Κώδικα Διευθέτησης Οφειλών και Παροχής Δεύτερης Ευκαιρίας, μεταξύ άλλων: 1ον. Θεσπίζονται διαδικασίες για την έγκαιρη προειδοποίηση των πολιτών στο πλαίσιο πρόληψης, ώστε να μην οδηγηθούν σε κατάσταση αφερεγγυότητας. 2ον. Εισάγεται ένα ολοκληρωμένο και αυτοματοποιημένο πλαίσιο ρύθμισης οφειλών, μέσω εξωδικαστικού μηχανισμού, τόσο για φυσικά όσο και για | The new framework for comprehensive debt settlement and second chance provision for households and businesses will be put out for public consultation within the day by the Ministry of Finance. The new framework aspires to be a holistic solution for the comprehensive settlement of private debt with quick and simple procedures by stakeholders. In a statement on the occasion of the publication of the bill, which will be posted and available for consultation in two weeks, Minister of Finance Christos Staikouras outlined the 12 key changes that the new framework brings. Specifically, the Finance Minister says that with the Debt Settlement and Second Chance Provision Code, among other thing: 1. Procedures are established for the early warning of citizens in the context of prevention, so that they are not led into insolvency. 2. A comprehensive and automated framework for debt settlement, through an out-of-court mechanism, is introduced for both natural and legal persons, so that |
| νομικά πρόσωπα, ώστε οι οφειλέτες να διατηρούν την περιουσία τους. 3ον. Δίνεται η δυνατότητα δεύτερης ευκαιρίας, με ταχείες διαδικασίες, με τη διαγραφή του υπολοίπου των οφειλών, κατόπιν ρευστοποίησης του συνόλου της περιουσίας και ελέγχου της περιουσιακής κατάστασης του οφειλέτη. 4ον. Ενσωματώνονται πρόνοιες για τον εντοπισμό των στρατηγικών κακοπληρωτών και πρόβλεψη συνεπειών για αυτούς. 5ον. Υπάρχουν πρόνοιες για δανειολήπτες που ανήκουν σε ευάλωτες κοινωνικά ομάδες, με την άσκηση επιδοματικής πολιτικής από το Κράτος, τόσο προληπτικά μέσω της επιδότησης δόσης δανείου, όσο και κατασταλτικά μέσω της επιδότησης ενοικίου. 6ον. Εκσυγχρονίζεται το πλαίσιο της διαδικασίας εξυγίανσης, ώστε να επιτρέψει στις επιχειρήσεις να αναδιαρθρώσουν επιτυχώς τις οφειλές τους και να επιστρέψουν σε παραγωγική λειτουργία, χωρίς να θίγονται τα δικαιώματα των εργαζομένων. 7ον. Απελευθερώνονται και αξιοποιούνται οι παραγωγικές μονάδες της χώρας που είναι δεσμευμένες σε ατέρμονες διαδικασίες ρύθμισης ή/και πτώχευσης, με σκοπό την επαναφορά τους σε λειτουργία, με την παράλληλη είσοδο εγχώριων και διεθνών επενδυτών. 8ον. Παρέχεται λύση στα μη εξυπηρετούμενα δάνεια, έτσι ώστε οι τράπεζες να μπορούν να συμβάλουν καθοριστικά στη χρηματοδότηση της οικονομίας. 9ον. Θεσπίζεται απαλλαγή των μελών διοίκησης του νομικού προσώπου που έχει πτωχεύσει, ώστε να μην εγκλωβίζονται για οφειλές που ανήκαν στην επιχείρηση. 10ον. Προάγεται η χρηματοοικονομική διαμεσολάβηση, που αποτελεί διεθνώς την πιο διαδεδομένη εξωδικαστική διαδικασία επίλυσης ιδιωτικών διαφορών στον χρηματοοικονομικό τομέα. 11ον. Αξιοποιείται η τεχνολογία, εισάγοντας νέες ηλεκτρονικές και αυτοματοποιημένες διαδικασίες, που διασφαλίζουν τη διαφάνεια, καταργούν τη γραφειοκρατία και προάγουν τον ψηφιακό μετασχηματισμό του Κράτους. 12ον. Εισάγονται ρυθμίσεις για τη βελτίωση του θεσμού των διαχειριστών αφερεγγυότητας, οι οποίοι καλούνται να διαδραματίσουν ουσιαστικό ρόλο στις διαδικασίες της πτώχευσης. Αυτές οι διαδικασίες, που εισάγονται με το νέο συνεκτικό και ολοκληρωμένο πλαίσιο διευθέτησης οφειλών και παροχής δεύτερης ευκαιρίας, αναμένεται να συμβάλουν στην ενίσχυση της ρευστότητας, της απασχόλησης και της κοινωνικής συνοχής, προστατεύοντας, στον μέγιστο δυνατό βαθμό, τον παραγωγικό ιστό της. | debtors retain their assets. 3. The possibility of a second chance, with fast-track procedures, is given, with the cancellation of the balance of the debts, following the liquidation of all the property and the verification of the debtor's financial situation. 4. Provisions are incorporated in order to identify strategic defaulters and provide for consequences for them. 5. There are provisions for borrowers belonging to vulnerable social groups, with the State exercising a subsidy policy, both preventively through the loan instalment subsidy and repressively through the rent subsidy. 6. Modernise the framework of the reorganisation process to allow enterprises to successfully restructure their debts and return to productive operation, without affecting the rights of workers. 7. The country's productive units that are tied up in endless regulation and/or bankruptcy proceedings are freed and utilized, with a view to bringing them back into operation, while allowing the entry of domestic and international investors. 8. A solution to non-performing loans is provided so that banks can make a decisive contribution to financing the economy. 9. An exemption is established for the management members of the bankrupt legal entity so that they are not trapped for debts that belonged to the company. 10. Financial mediation is promoted, which is the most widespread internationally used out-of-court procedure for the resolution of private disputes in the financial sector. 11. Technology is exploited by introducing new electronic and automated procedures that ensure transparency, eliminate bureaucracy and promote the digital transformation of the State. 12. Provisions are introduced to improve the institution of insolvency administrators, who are called upon to play an essential role in bankruptcy proceedings. These procedures, introduced by the new coherent and integrated framework for debt settlement and second chance, are expected to contribute to enhancing liquidity, employment and social cohesion, while protecting, to the maximum extent possible, the production infrastructure. |
| Human author summary | |
| Το νέο πλαίσιο φιλοδοξεί να αποτελέσει μια ολιστική λύση για τη συνολική διευθέτηση του ιδιωτικού χρέους με | The new framework aspires to be a holistic solution for the comprehensive settlement of private debt with quick |
| γρήγορες και απλές διαδικασίες από τους ενδιαφερόμενους. Ποιες είναι οι 12 βασικές αλλαγές. | and simple procedures by stakeholders. What are the 12 key changes. |
| Greek-mT5-small | |
| Σε δημόσια διαβούλευση τίθεται εντός της ημέρας το νέο πλαίσιο για την συνολική διευθέτηση οφειλών και την παροχή δεύτερης ευκαιρίας για νοικοκυριά και επιχειρήσεις. | The new framework for comprehensive debt settlement and second chance provision for households and businesses will be put out for public consultation within the day. |
| Greek-umT5-small | |
| Τι περιλαμβάνει το νέο πλαίσιο για την συνολική διευθέτηση οφειλών και την παροχή δεύτερης ευκαιρίας για νοικοκυριά και επιχειρήσεις. | What is included in the new framework for comprehensive debt settlement and second-chance provision for households and businesses. |
| Greek-umT5-base | |
| Τι περιλαμβάνει ο νέος Κώδικας Διευθέτησης Οφειλών και Παροχής Δεύτερης Ευκαιρίας για νοικοκυριά και επιχειρήσεις. | What is included in the new Debt Settlement and Second Chance Provision Code for households and businesses. |
| GreekBART | |
| Σε δημόσια διαβούλευση τίθεται εντός της ημέρας το νέο πλαίσιο για την συνολική διευθέτηση οφειλών και την παροχή δεύτερης ευκαιρίας για νοικοκυριά και επιχειρήσεις. | The new framework for comprehensive debt settlement and second chance provision for households and businesses will be put out for public consultation within the day. |
| TextRank | |
| Αυτές οι διαδικασίες, που εισάγονται με το νέο συνεκτικό και ολοκληρωμένο πλαίσιο διευθέτησης οφειλών και παροχής δεύτερης ευκαιρίας, αναμένεται να συμβάλουν στην ενίσχυση της ρευστότητας, της απασχόλησης και της κοινωνικής συνοχής, προστατεύοντας, στον μέγιστο δυνατό βαθμό, τον παραγωγικό ιστό της. | These procedures, introduced by the new coherent and integrated framework for debt settlement and second chance, are expected to contribute to enhancing liquidity, employment and social cohesion, while protecting, to the maximum extent possible, the production infrastructure. |

**Example 2**

| Article | |
|---|---|
| Για τέταρτη φορά θα τραγουδήσει σε σόου της Eurovision η Έλενα Παπαρίζου. Η τραγουδίστρια που χάρισε τη νίκη στην Ελλάδα το 2005 στο Κίεβο με το Number One θα είναι καλεσμένη και του Europe Shine a Light, του σόου που ετοιμάζει η EBU με τους καλλιτέχνες να παρουσιάζουν τα φετινά τραγούδια από το σπίτι τους, εξαιτίας του κορονοϊού. Με τη φετινή Eurovision να ακυρώνεται (θα γίνει το 2021 στο Ρότερνταμ), θα | For the fourth time Elena Paparizou will sing in a Eurovision show. The singer who gave Greece the victory in 2005 in Kiev with Number One will also be a guest of Europe Shine a Light, the show prepared by the EBU with the artists presenting this year's songs from their homes due to coronavirus. With this year's Eurovision cancelled (it will be held in Rotterdam in 2021), we will be watching a different show on Saturday 16 May, which will also be |
| παρακολουθήσουμε το Σάββατο 16 Μαΐου ένα διαφορετικό σόου, το οποίο θα μεταδώσει και η ΕΡΤ. Οι καλλιτέχνες που θα έπαιρναν φέτος μέρος στο θεσμό θα παρουσιάσουν τα τραγούδια τους, ενώ θα υπάρχουν και ειδικοί καλεσμένοι μέσω live κάμερας. Εκτός από την Παπαρίζου, όμως, θα δούμε και την αγαπημένη πλέον της διοργάνωσης Ελένη Φουρέιρα, η οποία δεν κατάφερε να πάρει τη νίκη με το 'Fuego', αλλά λατρεύτηκε από τους Eurofans. Δείτε το σχετικό απόσπασμα από την εκπομπή 'Στη Φωλιά των Κούκου' του Star:" | broadcast by ERT. The artists taking part in this year's event will perform their songs, and there will be special guests via live camera. Apart from Paparizou, however, we will also see the event's now favourite Eleni Foureira, who didn't manage to win with 'Fuego', but was adored by Eurofans. Watch the relevant excerpt from Star's show 'In the Cuckoo's Nest':" |
| Human author summary | |
| Η Έλενα Παπαρίζου θα τραγουδήσει live από το σπίτι της στο σόου που ετοιμάζει η Eurovision για τις 16 Μαΐου και θα μεταδώσει η ΕΡΤ, ενώ θα δούμε και την Ελένη Φουρέιρα. | Elena Paparizou will sing live from her home in the show that Eurovision is preparing for the 16th of May and will be broadcasted by ERT, and we will also see Eleni Foureira. |
| Greek-mT5-small | |
| Η Έλενα Παπαρίζου θα τραγουδήσει για τέταρτη φορά σε σόου της Eurovision. | Elena Paparizou will sing for the fourth time in a Eurovision show. |
| Greek-umT5-small | |
| Η Ελένα Παπαρίζου θα τραγουδήσει για τέταρτη φορά σε σόου της Eurovision, με τους καλλιτέχνες να παρουσιάζουν τα φετινά τραγούδια από το σπίτι τους, εξαιτίας του κορονοϊού. | Elena Paparizou will sing for the fourth time in a Eurovision show, with performers presenting this year's songs from their homes due to the coronavirus. |
| Greek-umT5-base | |
| Η τραγουδίστρια που χάρισε τη νίκη στην Ελλάδα το 2005 στο Κίεβο θα τραγουδήσει για τέταρτη φορά σε σόου της Eurovision. Δείτε το σχετικό απόσπασμα. | The singer who gave the victory to Greece in 2005 in Kiev will sing for the fourth time in a Eurovision show. Watch the relevant clip. |
| GreekBART | |
| Για τέταρτη φορά θα τραγουδήσει σε σόου της Eurovision η Έλενα Παπαρίζου. Θα δούμε και την Ελένη Φουρέιρα. | For the fourth time Elena Paparizou will sing in a Eurovision show. We will also see Eleni Foureira. |
| TextRank | |
| Η τραγουδίστρια που χάρισε τη νίκη στην Ελλάδα το 2005 στο Κίεβο με το Number One θα είναι καλεσμένη και του Europe Shine a Light, του σόου που ετοιμάζει η EBU με τους καλλιτέχνες να παρουσιάζουν τα φετινά τραγούδια από το σπίτι τους, εξαιτίας του κορονοϊού. | The singer who gave Greece the victory in 2005 in Kiev with Number One will also be a guest of Europe Shine a Light, the show prepared by the EBU with the artists presenting this year's songs from their homes due to coronavirus. |

**Example 3**

| Article | |
|---|---|
| Στα περίπου 317,1 δισ. ευρώ ή στο 177,1% του ΑΕΠ διαμορφώθηκε το δημόσιο χρέος στο τέλος του 2014, σύμφωνα με τα στοιχεία που δημοσιοποίησε η ΕΛΣΤΑΤ στη Eurostat. Παράλληλα, στο ισοζύγιο της Γενικής Κυβέρνησης καταγράφηκε έλλειμμα ύψους 6,4 δισ. ευρώ ή 3,5% του ΑΕΠ (το συγκεκριμένο μέγεθος ορίζεται στο πλαίσιο της διαδικασίας υπερβολικού ελλείμματος και δεν σχετίζεται με το δημοσιονομικό πλεόνασμα/έλλειμμα του Μνημονίου). Επίσης, σύμφωνα με τα ίδια στοιχεία της ΕΛΣΤΑΤ, το ΑΕΠ διαμορφώθηκε στο τέλος του 2014 στα 179,081 δισ. ευρώ, από 182,438 δισ. ευρώ το 2013 και 207,752 δισ. ευρώ στο τέλος του 2010. Το δημόσιο χρέος (εξαιρουμένου του έτους 2012, με το PSI), συνεχίζει να αυξάνεται ως ποσοστό του ΑΕΠ, αν και μειώνεται σε ονομαστική αξία. Ειδικότερα, διαμορφώθηκε σε 355,977 δισ. ευρώ (171,3% του ΑΕΠ) το 2010, σε 304,714 δισ. ευρώ (156,9% του ΑΕΠ) το 2011, σε 319,178 δισ. ευρώ (175% του ΑΕΠ) το 2013 και σε 317,094 δισ. ευρώ (177,1% του ΑΕΠ) το 2014. Σημειώνεται ότι, με βάση τα στοιχεία της ΕΛΣΤΑΤ, εάν συνυπολογισθούν οι δαπάνες για την κρατική υποστήριξη στα χρηματοπιστωτικά ιδρύματα της χώρας και η επίπτωσή τους στο ισοζύγιο της Γενικής Κυβέρνησης, μπορεί να προκύψει ως αποτέλεσμα πρωτογενές πλεόνασμα ύψους 0,4% του ΑΕΠ. (Με πληροφορίες από ΑΠΕ). | According to data published by ELSTAT to Eurostat, public debt stood at around 317.1 billion euros or 177.1% of GDP at the end of 2014, while the general government balance recorded a deficit of 6.4 billion euros or 3.5% of GDP (this figure is defined in the context of the excessive deficit procedure and is not related to the fiscal surplus/deficit of the Memorandum). Also, according to the same ELSTAT data, GDP stood at €179.081 billion at the end of 2014, down from €182.438 billion in 2013 and €207.752 billion at the end of 2010. Public debt (excluding the year 2012, with the PSI) continues to increase as a percentage of GDP, although it is decreasing in nominal terms. In particular, it stood at EUR 355.977 billion (171.3% of GDP) in 2010, EUR 304.714 billion (156.9% of GDP) in 2011, EUR 319.178 billion (175% of GDP) in 2013 and EUR 317.094 billion (177.1% of GDP) in 2014. It should be noted that, based on ELSTAT data, if the expenditure on state support to the country's financial institutions and its impact on the general government balance are taken into account, it may result in a primary surplus of 0.4% of GDP. (With information from APE). |
| **Human author summary** | |
| Στο 171,1% του ΑΕΠ το δημόσιο χρέος της χώρας στο τέλος του 2014. Πόσο αντιστοιχεί το ποσοστό σε απόλυτους αριθμούς. Αναλυτικά τα στοιχεία που έδωσε η ΕΛΣΤΑΤ στη Eurostat. | The country's public debt stood at 171.1% of GDP at the end of 2014. What is the percentage in absolute terms. Details of the data provided by ELSTAT to Eurostat. |
| **Greek-mT5-small** | |
| Τι δείχνουν τα στοιχεία της ΕΛΣΤΑΤ για το δημόσιο χρέος. | What do the ELSTAT data show on public debt. |
| **Greek-umT5-small** | |
| Στο ισοζύγιο της Γενικής Κυβέρνησης καταγράφηκε έλλειμμα ύψους 6,4 δισ. ευρώ ή 3,5% του ΑΕΠ. | The general government balance recorded a deficit of 6.4 billion euros or 3.5% of GDP. |
| **Greek-umT5-base** | |
| Τι δείχνουν τα στοιχεία της Eurostat για το δημόσιο χρέος. Υπερπλεόνασμα 6,4 δισ. ευρώ στο ισοζύγιο της Γενικής Κυβέρνησης. | What do the Eurostat data on public debt show. A surplus of €6.4 billion in the general government balance. |
| GreekBART | |
| Στα 317,1 δισ. ευρώ ή στο 177,1 του ΑΕΠ διαμορφώθηκε το δημόσιο χρέος στο τέλος του 2014. Τι δείχνουν τα στοιχεία της ΕΛΣΤΑΤ. | The public debt stood at 317.1 billion euros or 177.1 of GDP at the end of 2014. What do the data from ELSTAT show. |
| TextRank | |
| Ειδικότερα, διαμορφώθηκε σε 355,977 δισ. ευρώ (171,3% του ΑΕΠ) το 2010, σε 304,714 δισ. ευρώ (156,9% του ΑΕΠ) το 2011, σε 319,178 δισ. ευρώ (175% του ΑΕΠ) το 2013 και σε 317,094 δισ. ευρώ (177,1% του ΑΕΠ) το 2014. | In particular, it stood at EUR 355.977 billion (171.3% of GDP) in 2010, EUR 304.714 billion (156.9% of GDP) in 2011, EUR 319.178 billion (175% of GDP) in 2013 and EUR 317.094 billion (177.1% of GDP) in 2014. |

**Example 4**

| Article | |
|---|---|
| Να πάρει ""ξεκάθαρη"" θέση σε σχέση με τον κίνδυνο μετάδοσης του κορονοϊού από τη Θεία Κοινωνία καλεί την κυβέρνηση και τον Πρωθυπουργό με ανακοίνωσή του τη Δευτέρα ο ΣΥΡΙΖΑ. ""Την ώρα που κλείνουν προληπτικά και ορθώς σχολεία, πανεπιστήμια, γήπεδα και λαμβάνονται ειδικά μέτρα ακόμη και για την ορκωμοσία της νέας Προέδρου της Δημοκρατίας, η Ιερά Σύνοδος της Εκκλησίας της Ελλάδος επιμένει ότι το μυστήριο της Θείας Κοινωνίας δεν εγκυμονεί κινδύνους μετάδοσης του κορονοϊού, καλώντας όμως τις ευπαθείς ομάδες να μείνουν σπίτι τους"", αναφέρει η αξιωματική αντιπολίτευση και συνεχίζει: ""Ωστόσο το πρόβλημα δεν είναι τι λέει η Ιερά Σύνοδος, αλλά τι λέει η Πολιτεία και συγκεκριμένα ο ΕΟΔΥ και το Υπουργείο Υγείας, που έχουν και την αποκλειστική κοινωνική ευθύνη για τη μη εξάπλωση του ιού και την προστασία των πολιτών"". ""Σε άλλες ευρωπαϊκές χώρες με εξίσου μεγάλο σεβασμό στη Χριστιανική πίστη και στο θρησκευτικό συναίσθημα, τα μυστήρια της Εκκλησίας είτε αναστέλλονται είτε τροποποιούν το τελετουργικό τους. Μόνο στη χώρα μας έχουμε το θλιβερό προνόμιο μιας πολιτείας που δεν τολμά να πει το αυτονόητο"", προσθέτει, τονίζοντας ότι ""η κυβέρνηση λοιπόν και το Υπουργείο Υγείας οφείλουν να πάρουν δημόσια μια ξεκάθαρη θέση και να μην θυσιάζουν τη δημόσια Υγεία στο βωμό του πολιτικού κόστους"". ""Συμφωνούν ότι η Θεία Κοινωνία δεν εγκυμονεί κινδύνους μετάδοσης του κορονοϊού; Δεν είναι θέμα | SYRIZA calls on the government and the Prime Minister to take a ""clear"" position on the risk of coronavirus transmission through Holy Communion in a statement on Monday. ""At a time when schools, universities and stadiums are being closed as a precautionary measure and rightly so, and special measures are being taken even for the swearing-in of the new President of the Republic, the Holy Synod of the Church of Greece insists that the sacrament of Holy Communion poses no risk of coronavirus transmission, while calling on vulnerable groups to stay at home,"" the opposition party also said: ""However, the problem is not what the Holy Synod says, but what the State says, and in particular the Hellenic Organization for Public Health (EODY) and the Public Health Ministry, which have the exclusive social responsibility for the non-spread of the virus and the protection of citizens."" ""In other European countries with equal respect for the Christian faith and religious sentiment, the sacraments of the Church are either suspended or their rituals are modified. Only in our country we have the sad privilege of a state that does not dare to say the obvious,"" they add, stressing that ""the government and the Ministry of Health must therefore take a clear public position and not sacrifice public health on the altar of political costs"". "Do they agree that Holy Communion does not pose any risk of coronavirus transmission? It is not a matter of piety but of social |
| ευσέβειας αλλά κοινωνικής ευθύνης. Και με τη Δημόσια υγεία δεν μπορούμε να παίζουμε"", καταλήγει η ανακοίνωση του γραφείου Τύπου του ΣΥΡΙΖΑ. *ΠΩΣ ΜΕΤΑΔΙΔΕΤΑΙ. Χρήσιμος οδηγός για να προστατευθείτε από τον κορονοϊό *ΤΑ ΝΟΣΟΚΟΜΕΙΑ ΑΝΑΦΟΡΑΣ. Ποια θα υποδέχονται τα κρούσματα κορονοϊού στην Ελλάδα. *ΤΑΞΙΔΙΑ. Κορονοϊός και αεροδρόμια: Τι να προσέξετε. *Η ΕΠΙΔΗΜΙΑ ΣΤΟΝ ΠΛΑΝΗΤΗ. Δείτε LIVE χάρτη με την εξέλιξη του κορονοϊού." | responsibility. And we cannot fool around with public health," concludes the statement from the Syriza press office. *HOW THE VIRUS IS TRANSMITTED. Useful guide to protect yourself from coronavirus * REFERENCE HOSPITALS. Which will receive coronavirus cases in Greece. *TRAVELLING. Coronavirus and airports: What to watch out for. * THE PLANET EPIDEMIC. See a LIVE map of coronavirus spreading." |
| Human author summary | |
| Ο ΣΥΡΙΖΑ κάλεσε τον Κυριάκο Μητσοτάκη να πάρει δημοσίως θέση για το αν συμφωνεί με την οδηγία της Ιεράς Συνόδου. | SYRIZA called on Kyriakos Mitsotakis to take a public position on whether he agrees with the directive of the Holy Synod. |
| Greek-mT5-small | |
| "Η κυβέρνηση λοιπόν και το Υπουργείο Υγείας οφείλουν να πάρουν δημόσια μια ξεκάθαρη θέση και να μην θυσιάζουν τη δημόσια Υγεία στο βωμό του πολιτικού κόστους" αναφέρει σε ανακοίνωσή του ο ΣΥΡΙΖΑ. | "The government and the Ministry of Health should take a clear public position and not sacrifice public health care on the altar of political costs," SYRIZA mentioned in a statement. |
| Greek-umT5-small | |
| "Η κυβέρνηση λοιπόν και το Υπουργείο Υγείας οφείλουν να πάρουν δημόσια μια ξεκάθαρη θέση και να μην θυσιάζουν τη δημόσια Υγεία στο βωμό του πολιτικού κόστους" αναφέρει ο ΣΥΡΙΖΑ. | "The government and the Ministry of Health should take a clear public position and not sacrifice public health care on the altar of political costs," SYRIZA mentions. |
| Greek-umT5-base | |
| "Η κυβέρνηση λοιπόν και το Υπουργείο Υγείας οφείλουν να πάρουν δημόσια μια ξεκάθαρη θέση και να μην θυσιάζουν τη δημόσια Υγεία στο βωμό του πολιτικού κόστους" αναφέρει ο ΣΥΡΙΖΑ. | "The government and the Ministry of Health should take a clear public position and not sacrifice public health care on the altar of political costs," SYRIZA mentions. |
| GreekBART | |
| Η Ιερά Σύνοδος της Εκκλησίας της Ελλάδος επιμένει ότι το μυστήριο της Θείας Κοινωνίας δεν εγκυμονεί κινδύνους μετάδοσης του κορονοϊού, καλώντας όμως τις ευπαθείς ομάδες να μείνουν σπίτι τους. | The Holy Synod of the Church of Greece insists that the sacrament of Holy Communion poses no risk of coronavirus transmission but urges vulnerable groups to stay at home. |
| TextRank | |
| ""Την ώρα που κλείνουν προληπτικά και ορθώς σχολεία, πανεπιστήμια, γήπεδα και λαμβάνονται ειδικά μέτρα ακόμη και για την ορκωμοσία της νέας Προέδρου της Δημοκρατίας, η Ιερά Σύνοδος της Εκκλησίας της Ελλάδος | “"At a time when schools, universities and stadiums are closed as a precautionary measure and special measures are taken even for the inauguration of the new President of the Republic, the Holy Synod of the Church of Greece |
| επιμένει ότι το μυστήριο της Θείας Κοινωνίας δεν εγκυμονεί κινδύνους μετάδοσης του κορονοϊού, καλώντας όμως τις ευπαθείς ομάδες να μείνουν σπίτι τους"", αναφέρει η αξιωματική αντιπολίτευση και συνεχίζει: ""Ωστόσο το πρόβλημα δεν είναι τι λέει η Ιερά Σύνοδος, αλλά τι λέει η Πολιτεία και συγκεκριμένα ο ΕΟΔΥ και το Υπουργείο Υγείας, που έχουν και την αποκλειστική κοινωνική ευθύνη για τη μη εξάπλωση του ιού και την προστασία των πολιτών"" | insists that the sacrament of Holy Communion does not pose any risk of coronavirus transmission, while calling on vulnerable groups to stay at home"", the opposition party says and continues: ""However, the problem is not what the Holy Synod says, but what the State says, and specifically the EODY and the Ministry of Health, which have the exclusive social responsibility for the non-spread of the virus and the protection of citizens"" |

**Example 5**

| Article | |
|---|---|
| Μέ άρθρο της με τίτλο ""Επιστρέψτε στη θεά Ίριδα το σώμα της"", η εφημερίδα Washington Post τάσσεται υπέρ της επιστροφής των γλυπτών του Παρθενώνα, στην Αθήνα, στην κοιτίδα του δυτικού πολιτισμού, τώρα που οι συνθήκες έχουν αλλάξει για την πάλαι ποτέ αυτοκρατορία της Αγγλίας. Αναφερόμενη στις διαφορετικές απόψεις Ελλήνων και Βρετανών για τα γλυπτά, η συντάκτρια του άρθρου, τονίζει ότι το αίτημα επιστροφής έχει αποκτήσει μεγαλύτερο βάρος τώρα που το Ηνωμένο Βασίλειο εγκαταλείπει την Ευρωπαϊκή Ένωση. «Όταν ο Τόμας Μπρους, έβδομος κόμης του Έλγιν, και 11ος κόμης του Κινκαρντίν, ταξίδεψε στην Ακρόπολη στις αρχές της δεκαετίας του 1800, ως Βρετανός πρέσβης στην Οθωμανική Αυτοκρατορία, ο Σουλτάνος λέγεται ότι του έδωσε την άδεια να ""αφαιρέσει μερικά τμήματα λίθων με παλιές επιγραφές και μορφές"". Ο λόρδος το εξέλαβε ως άδεια να αφαιρέσει, περίπου, 17 αγάλματα από τα αετώματα, 15 μετώπες, και 247 πόδια (περίπου 75 μέτρα) της ζωφόρου από τον Παρθενώνα για να τα φέρει στην καλή μας Αγγλία» αναφέρει στο άρθρο της η Washington Post. Και συνεχίζει λέγοντας ότι «οι καιροί όμως άλλαξαν και αυτό που θεωρούνταν πιο δικαιολογημένο τότε, σήμερα θεωρείται ευρέως ως μια ασυνείδητη πράξη». Σε μία έμμεση αναφορά στο Brexit, και υπεραμυνόμενη της επιστροφής των γλυπτών στην Ελλάδα, η συντάκτρια του άρθρου της Washington Post, διερωτάται: «Γιατί να παραμείνουν τα μάρμαρα στη φύλαξη της χώρας που επιμένει ότι ανήκει μόνο στον εαυτό της;» και σημειώνει: «Η Ελλάδα τιμάται σήμερα ως λίκνο του δυτικού πολιτισμού, και ποιοί παρά οι Έλληνες θα μπορούσαν να στεγάσουν τον πολιτισμό αυτό;». | In an article entitled "Return to the goddess Iris her body", the Washington Post advocates the return of the Parthenon sculptures to Athens, the cradle of Western civilization, now that circumstances have changed for the former empire of England. Referring to the differing views of Greeks and Britons on the sculptures, the author of the article points out that the demand for their return has gained more weight now that the United Kingdom is leaving the European Union. "When Thomas Bruce, seventh Earl of Elgin, and 11th Earl of Kincardin, travelled to the Acropolis in the early 1800s as British ambassador to the Ottoman Empire, the Sultan is said to have given him permission to 'remove some sections of stones with old inscriptions and figures'. The lord took it as permission to remove, roughly, 17 statues from the pediments, 15 fronts, and 247 feet (about 75 meters) of frieze from the Parthenon to bring them to our good England," the Washington Post said in its article. It goes on to say that "times have changed, however, and what was considered more justifiable then is now widely seen as an unconscionable act." In an implicit reference to Brexit, and advocating for the return of the sculptures to Greece, the author of the Washington Post article asks, "Why should the marbles remain in the custody of the country that insists it belongs only to itself?" and notes: "Greece is now revered as the cradle of Western civilization, and to whom other than Greeks could this civilization be housed?" |
| Human author summary | |
| Η αμερικανική εφημερίδα τονίζει σε άρθρο της ότι το αίτημα επιστροφής έχει αποκτήσει μεγαλύτερο βάρος τώρα που το Ηνωμένο Βασίλειο εγκαταλείπει την Ευρωπαϊκή Ένωση. | The American newspaper points out in an article that the demand for return has gained more weight now that the United Kingdom is leaving the European Union. |
| Greek-mT5-small | |
| Η Washington Post τάσσεται υπέρ της επιστροφής των γλυπτών του Παρθενώνα στην Αθήνα, στην κοιτίδα του δυτικού πολιτισμού, τώρα που οι συνθήκες έχουν αλλάξει για την πάλαι ποτέ αυτοκρατορία της Αγγλίας. | The Washington Post argues in favor of returning the Parthenon sculptures to Athens, the cradle of Western civilization, now that circumstances have changed for the former empire of England. |
| Greek-umT5-small | |
| Η Washington Post τάσσεται υπέρ της επιστροφής των γλυπτών του Παρθενώνα στην Αθήνα, στην κοιτίδα του δυτικού πολιτισμού, τώρα που οι συνθήκες έχουν αλλάξει για την πάλαι ποτέ αυτοκρατορία της Αγγλίας. | The Washington Post argues in favor of returning the Parthenon sculptures to Athens, the cradle of Western civilization, now that circumstances have changed for the former empire of England. |
| Greek-umT5-base | |
| Υπέρ της επιστροφής των γλυπτών του Παρθενώνα, στην Αθήνα, στην κοιτίδα του δυτικού πολιτισμού τάσσεται η Washington Post. | In favour of returning the Parthenon sculptures to Athens, the cradle of Western civilization, argues the Washington Post. |
| GreekBART | |
| Με άρθρο της η εφημερίδα Washington Post τάσσεται υπέρ της επιστροφής των γλυπτών του Παρθενώνα, στην Αθήνα. | The Washington Post has an article in favour of the return of the Parthenon sculptures to Athens. |
| TextRank | |
| Σε μία έμμεση αναφορά στο Brexit, και υπεραμυνόμενη της επιστροφής των γλυπτών στην Ελλάδα, η συντάκτρια του άρθρου της Washington Post, διερωτάται: «Γιατί να παραμείνουν τα μάρμαρα στη φύλαξη της χώρας που επιμένει ότι ανήκει μόνο στον εαυτό της;» και σημειώνει: «Η Ελλάδα τιμάται σήμερα ως λίκνο του δυτικού πολιτισμού, και ποιοί παρά οι Έλληνες θα μπορούσαν να στεγάσουν τον πολιτισμό αυτό;». | In an implicit reference to Brexit, and advocating for the return of the sculptures to Greece, the author of the Washington Post article asks, "Why should the marbles remain in the custody of the country that insists it belongs only to itself?" and notes: "Greece is now revered as the cradle of Western civilization, and to whom other than Greeks could this civilization be housed?" |

**Example 6**

| Article | |
|---|---|
| «Οι έρευνες κοινής γνώμης, και ιδιαιτέρως οι δημοσκοπήσεις, είναι ένα εξαιρετικά χρήσιμο εργαλείο, αφού αποτυπώνουν τις τάσεις της κοινωνίας σε συγκεκριμένο χρόνο» σημειώνει ο Σύλλογος Εταιρειών | "Public opinion surveys, and especially opinion polls, are an extremely useful tool, since they reflect the trends of society at a specific time", notes the Association of Polling and Market Research Companies (SEDEA), in response to |
| Δημοσκόπησης και Έρευνας Αγοράς (ΣΕΔΕΑ) με αφορμή όσα ακολούθησαν σχετικά με τα αποτελέσματα των δημοσκοπήσεων και του exit poll στις εκλογές. Σύμφωνα με ανακοίνωση που εξεδόθη «οι δημοσκοπήσεις δεν δίνουν το αποτέλεσμα των εκλογών. Αποτυπώνουν την στιγμή και κάθε στιγμή είναι διαφορετική, αφού επηρεάζεται από πληθώρα γεγονότων που συμβαίνουν καθημερινά, τόσο κατά την προεκλογική περίοδο όσο και την περίοδο που προηγείται αυτής, όπως οι ίδιοι οι πολίτες μπορούν να διαπιστώσουν ανατρέχοντας στα αποτελέσματα των τελευταίων μηνών». «Εδώ και πολλά χρόνια, το εργαλείο αυτό αμφισβητείται έντονα με δημόσιες τοποθετήσεις από συγκεκριμένες πολιτικές ομάδες, με στόχο την αποδόμησή του. Η πραγματικότητα κατέδειξε, όχι μόνον στις παρούσες εκλογές αλλά και στις προηγούμενες, ότι όποιος την αρνείται αυτοπαγιδεύεται στο δικό του ιστόρημα. Οι φετινές εκλογές κατέδειξαν για άλλη μια φορά ότι το EXIT POLL αποτύπωσε σωστά την εκλογική συμπεριφορά, ενώ οι δημοσκοπήσεις καταγράφουν ικανοποιητικά τις τάσεις της κοινωνίας και μάλιστα σε περιόδους αστάθειας, σε ευρωπαϊκό και παγκόσμιο επίπεδο, όπου συντελούνται δομικές κοινωνικές αλλαγές. Η μόνη απόκλιση στο 100% του exit poll από τα τελικά αποτελέσματα των εκλογών αφορούσε στο ποσοστό του ΣΥΡΙΖΑ» προσθέτει ο ΣΕΔΕΑ και συμπληρώνει: «Σχετικά με τις αποκλίσεις των προεκλογικών δημοσκοπήσεων αλλά και του exit poll, η πρώτη μας εκτίμηση είναι ότι αυτές οφείλονται: Ευχή μας είναι να αφεθεί εφεξής ο κλάδος απερίσπαστος να κάνει την δουλειά του με την ευσυνειδησία και την ανεξαρτησία που διακρίνει τα μέλη του ΣΕΔΕΑ και με γνώμονα ότι οι δημοσκοπήσεις δεν είναι το αποτέλεσμα των εκλογών, αλλά η καταγραφή των τάσεων της κοινωνίας στο δρόμο προς τις εκλογές» καταλήγει. Ακολουθήστε το News 24/7 στο Google News και μάθετε πρώτοι όλες τις ειδήσεις." | the events that followed regarding the results of opinion polls and exit polls in the elections. According to a statement issued, "polls do not give the outcome of the elections. They reflect the moment, and each moment is different, since it is influenced by a multitude of events that happen every day, both during the election period and the period preceding it, as citizens themselves can see by looking back at the results of the last few months". "For many years now, this tool has been strongly contested by public statements by certain political groups with the aim of deconstructing it. Reality has shown, not only in this election but also in previous ones, that anyone who denies it is deluding himself in his own web. This year's elections have shown once again that the EXIT POLL has correctly captured electoral behaviour, while the polls are a satisfactory record of trends in society, even in times of instability, at European and global level, when structural social changes are taking place. The only deviation in 100% of the exit poll from the final results of the elections concerned the percentage of SYRIZA" adds SEDEA and adds: "Regarding the deviations of the pre-election polls and the exit poll, our first assessment is that they are due to: It is our wish that henceforth the industry be left uninhibited to do its work with the conscientiousness and independence that distinguishes the members of SEDEA and with the understanding that the polls are not the result of the elections, but the recording of the trends of society on the road to the elections", it concludes. Follow News 24/7 on Google News and be the first to know all the news." |
| Human author summary | |
| Με ανακοίνωσή του, ο Σύλλογος Εταιρειών Δημοσκόπησης και Έρευνας Αγοράς (ΣΕΔΕΑ) τοποθετείται επί του εκλογικού αποτελέσματος. | In a statement, the Association of Polling and Market Research Companies (SEDEA) commented on the election result. |
| Greek-mT5-small | |
| Οι δημοσκοπήσεις δεν δίνουν το αποτέλεσμα των εκλογών. Αποτυπώνουν την στιγμή και κάθε στιγμή είναι διαφορετική, αφού επηρεάζεται από πληθώρα γεγονότων | The polls do not give the result of the elections. They capture the moment and each moment is different, as it is influenced by a multitude of events that happen every day," |
| που συμβαίνουν καθημερινά" αναφέρει ο Σύλλογος Εταιρειών Δημοσκόπησης και Έρευνας Αγοράς. | says the Association of Polling and Market Research Companies. |
| Greek-umT5-small | |
| "Οι δημοσκοπήσεις δεν δίνουν το αποτέλεσμα των εκλογών. Αποτυπώνουν την στιγμή και κάθε στιγμή είναι διαφορετική, αφού επηρεάζεται από πληθώρα γεγονότων που συμβαίνουν καθημερινά" αναφέρει ο Σύλλογος Εταιρειών Δημοσκόπησης και Έρευνας Αγοράς. | "The polls do not give the result of the elections. They capture the moment and each moment is different, as it is affected by a multitude of events that happen every day," says the Association of Polling and Market Research Companies. |
| Greek-umT5-base | |
| "Οι δημοσκοπήσεις δεν δίνουν το αποτέλεσμα των εκλογών" αναφέρει σε ανακοίνωσή του ο Σύλλογος Εταιρειών Δημοσκόπησης και Έρευνας Αγοράς. | "Polls do not give the result of the elections," the Association of Polling and Market Research Companies said in a statement. |
| GreekBART | |
| "Οι έρευνες κοινής γνώμης, και ιδιαιτέρως οι δημοσκοπήσεις, είναι ένα εξαιρετικά χρήσιμο εργαλείο, αφού αποτυπώνουν τις τάσεις της κοινωνίας σε συγκεκριμένο χρόνο" σημειώνει ο Σύλλογος Εταιρειών Δημοσκόπησης και Έρευνας Αγοράς. | "Public opinion surveys, and in particular opinion polls, are an extremely useful tool, since they reflect the trends of society at a specific time", notes the Association of Polling and Market Research Companies. |
| TextRank | |
| Η μόνη απόκλιση στο 100% του exit poll από τα τελικά αποτελέσματα των εκλογών αφορούσε στο ποσοστό του ΣΥΡΙΖΑ» προσθέτει ο ΣΕΔΕΑ και συμπληρώνει: «Σχετικά με τις αποκλίσεις των προεκλογικών δημοσκοπήσεων αλλά και του exit poll, η πρώτη μας εκτίμηση είναι ότι αυτές οφείλονται: Ευχή μας είναι να αφεθεί εφεξής ο κλάδος απερίσπαστος να κάνει την δουλειά του με την ευσυνειδησία και την ανεξαρτησία που διακρίνει τα μέλη του ΣΕΔΕΑ και με γνώμονα ότι οι δημοσκοπήσεις δεν είναι το αποτέλεσμα των εκλογών, αλλά η καταγραφή των τάσεων της κοινωνίας στο δρόμο προς τις εκλογές» καταλήγει. | The only deviation in 100% of the exit poll from the final results of the elections concerned the percentage of SYRIZA" adds SEDEA and adds: "Regarding the deviations of the pre-election polls and the exit poll, our first assessment is that they are due to: It is our wish that henceforth the industry be left uninhibited to do its work with the conscientiousness and independence that distinguishes the members of SEDEA and with the understanding that the polls are not the result of the elections, but the recording of the trends of society on the road to the elections", it concludes. |

**Example 7**

| Article | |
|---|---|
| Αποδεκτή έκανε ο ΣΥΡΙΖΑ, δια στόματος Πόπης Τσαπανίδου την πρόταση του MEGA για διεξαγωγή ντιμπέιτ μεταξύ του Αλέξη Τσίπρα και του Κυριάκου Μητσοτάκη. ""Με χαρά αποδεχόμαστε την πρόσκληση του τηλεοπτικού σταθμού MEGA σε ανοιχτό διάλογο | SYRIZA, through the mouth of Popi Tsapanidou, accepted MEGA's proposal for a debate between Alexis Tsipras and Kyriakos Mitsotakis. "We are pleased to accept the invitation of the MEGA television station to hold an open debate between Alexis Tsipras and Kyriakos Mitsotakis. |
| ντιμπέιτ μεταξύ του Αλέξη Τσίπρα και του Κυριάκου Μητσοτάκη. Είμαστε στη διάθεση του καναλιού, όποια μέρα επιθυμεί τροποποιώντας το πρόγραμμα του προέδρου για να γίνει ο δημόσιος διάλογος που θα συμβάλει στην ενημέρωση των πολιτών"", αναφέρει η δήλωση της Πόπης Τσαπανίδου. Υπενθυμίζεται ότι η σχετική πρόταση έγινε από τον Γενικό Διευθυντή Ειδήσεων του καναλιού Σταμάτη Μαλέλη, καθώς ο τηλεοπτικός σταθμός MEGA δεν θα μπορέσει να φιλοξενήσει την τηλεμαχία με όλους τους αρχηγούς, λόγω της προβολής ποδοσφαιρικού αγώνα. Στην πρόσκλησή του ο κ. Μαλέλης ανέφερε χαρακτηριστικά: ""Θεωρούμε ότι η ανταλλαγή απόψεων Μητσοτάκη - Τσίπρα θα συμβάλει στην πληρέστερη ενημέρωση των πολιτών ενόψει των κρίσιμων εκλογών της 21ης Μαΐου"". "Όχι" σε debate μεταξύ του Κυριάκου Μητσοτάκη και του Αλέξη Τσίπρα στο Mega λέει το Μαξίμου, απαντώντας στο ΣΥΡΙΖΑ. Ειδικότερα με δήλωση του ο υπουργός Επικρατείας και κυβερνητικός εκπρόσωπος, Άκης Σκέρτσος, απάντησε στην εκπρόσωπο Τύπου του ΣΥΡΙΖΑ Πόπη Τσαπανίδου: "Η διακομματική επιτροπή, δηλαδή ο συναινετικός θεσμός που εδώ και δεκαετίες ορίζει τις διαδικασίες και τους όρους διεξαγωγής του προεκλογικού αγώνα, αποφάσισε για ένα debate μεταξύ των έξι επικεφαλής των κοινοβουλευτικών κομμάτων την Τετάρτη 10 Μαΐου. Αυτό αποφασίσθηκε ομόφωνα και αυτό θα γίνει. Ο ΣΥΡΙΖΑ μπορεί να προσκαλεί ή να αυτοπροσκαλείται, να επιβεβαιώνει ότι είναι εξαίρεση όσο θέλει. Η δυσανεξία του στους θεσμούς δεν μας εκπλήσσει. Η στάση του είναι αποκαλυπτική ως προς τον σεβασμό του στα υπόλοιπα κόμματα της αντιπολίτευσης. Η απομόνωσή του είναι καθολική και η στάση του κωμική."΄ Ακολουθήστε το News 24/7 στο Google News και μάθετε πρώτοι όλες τις ειδήσεις. | We are at the channel's disposal, any day it wishes by modifying the president's program in order to have a public dialogue that will contribute to informing citizens," reads the statement of Popi Tsapanidou. It is recalled that the proposal was made by the General Director of News of the channel Stamatis Malelis, as the TV station MEGA will not be able to host the tele-conference with all the leaders, due to the showing of a football match. In his invitation, Mr.Malelis said: "We believe that the exchange of views between Mitsotakis and Tsipras will contribute to the fuller information of citizens in view of the crucial elections of May 21". "No" to a debate between Kyriakos Mitsotakis and Alexis Tsipras on Mega, says Maximou, responding to SYRIZA. In particular, State Minister and government spokesman Akis Skertosos responded to SYRIZA's press spokesperson Poppi Tsapanidou in a statement: "The bipartisan committee, that is, the consensual institution that for decades has defined the procedures and conditions of the election campaign, decided on a debate between the six heads of the parliamentary parties on Wednesday, May 10. This was decided unanimously and that is what will happen. Syriza can invite or self-invite, confirm that it is an exception as much as it wants. Its intolerance of the institutions does not surprise us. Its attitude is revealing as to its respect for the other opposition parties. His isolationism is universal and his attitude is comical."΄ Follow News 24/7 on Google News and be the first to know all the news. |
| Human author summary | |
| Την πρόταση του MEGA για ντιμπέιτ μεταξύ του Αλέξη Τσίπρα και του Κυριάκου Μητσοτάκη, αποδέχθηκε ο ΣΥΡΙΖΑ. Τι δήλωσε η Πόπη Τσαπανίδου. | The proposal of MEGA for a debate between Alexis Tsipras and Kyriakos Mitsotakis was accepted by SYRIZA. What did Poppi Tsapanidou state. |
| Greek-mT5-small | |
| Αποδεκτή έκανε ο ΣΥΡΙΖΑ την πρόταση του MEGA για διεξαγωγή ντιμπέιτ μεταξύ του Αλέξη Τσίπρα και του Κυριάκου Μητσοτάκη. | SYRIZA accepted MEGA's proposal for a debate between Alexis Tsipras and Kyriakos Mitsotakis. |
| Greek-umT5-small | |
| Η πρόταση του MEGA για διεξαγωγή ντιμπέιτ μεταξύ του Αλέξη Τσίπρα και του Κυριάκου Μητσοτάκη έγινε από τον Γενικό Διευθυντή Ειδήσεων του καναλιού Σταμάτη Μαλέλη. | MEGA's proposal for a debate between Alexis Tsipras and Kyriakos Mitsotakis was made by the channel's General Director of News Stamatis Malelis. |
| Greek-umT5-base | |
| Η πρόταση του MEGA για διεξαγωγή ντιμπέιτ μεταξύ του Αλέξη Τσίπρα και του Κυριάκου Μητσοτάκη, αποδέχτηκε ο ΣΥΡΙΖΑ, δια στόματος Πόπης Τσαπανίδου. | MEGA's proposal for a debate between Alexis Tsipras and Kyriakos Mitsotakis was accepted by SYRIZA, through the mouth of Popi Tsapanidou. |
| GreekBART | |
| "Είμαστε στη διάθεση του MEGA για να γίνει ο δημόσιος διάλογος που θα συμβάλει στην ενημέρωση των πολιτών" δήλωσε η Πόπη Τσαπανίδου, απαντώντας στην πρόταση του MEGA. | "We are at MEGA's disposal for a public dialogue that will contribute to informing citizens," said Poppy Tsapanidou, responding to MEGA's proposal. |
| TextRank | |
| Ειδικότερα με δήλωση του ο υπουργός Επικρατείας και κυβερνητικός εκπρόσωπος, Άκης Σκέρτσος, απάντησε στην εκπρόσωπο Τύπου του ΣΥΡΙΖΑ Πόπη Τσαπανίδου: "Η διακομματική επιτροπή, δηλαδή ο συναινετικός θεσμός που εδώ και δεκαετίες ορίζει τις διαδικασίες και τους όρους διεξαγωγής του προεκλογικού αγώνα, αποφάσισε για ένα debate μεταξύ των έξι επικεφαλής των κοινοβουλευτικών κομμάτων την Τετάρτη 10 Μαΐου. | In particular, State Minister and government spokesman Akis Skertosos responded to SYRIZA's press spokesperson Poppi Tsapanidou in a statement: "The bipartisan committee, that is, the consensual institution that for decades has defined the procedures and conditions of the election campaign, decided on a debate between the six heads of the parliamentary parties on Wednesday, May 10. |

**Example 8**

| | |
|---|---|
| Article | |
| Το Υπουργείο Εργασίας, Κοινωνικής Ασφάλισης και Κοινωνικής Αλληλεγγύης, ενεργοποίησε το εργαλείο της εφάπαξ οικονομικής ενίσχυσης ύψους χιλίων ευρώ, προς χιλιάδες ανέργους και εργαζόμενους σε επίσχεση οι οποίοι επλήγησαν ιδιαίτερα από τις συνέπειες της οικονομικής κρίσης. Σύμφωνα με τη σχετική ανακοίνωση, η χορήγηση της ενίσχυσης αυτής είχε ουσιαστικά ανασταλεί την περίοδο 2012-2014, καθώς η προηγούμενη κυβέρνηση ενέκρινε μόνο ένα σχετικό αίτημα σωματείου. Κατά την τριετία 2015-2017 από την κυβέρνηση ΣΥΡΙΖΑ/ΑΝΕΛ εγκρίθηκαν συνολικά 50 σχετικά αιτήματα. Η Υπουργός Εργασίας, Κοινωνικής Ασφάλισης και Κοινωνικής Αλληλεγγύης, Έφη Αχτσιόγλου, υπέγραψε 12 νέες αποφάσεις για τη χορήγηση εφάπαξ οικονομικής ενίσχυσης ύψους χιλίων | The Ministry of Labour, Social Security and Social Solidarity has activated the tool of one-off financial support of one thousand euros to thousands of unemployed and retrenched workers who have been particularly affected by the consequences of the economic crisis. According to the announcement, the granting of this aid was effectively suspended in the period 2012-2014, as the previous government approved only one union request. In the three-year period 2015-2017, a total of 50 relevant requests were approved by the SYRIZA/ANEL government. The Minister of Labour, Social Security and Social Solidarity, Efi Achtsioglou, signed 12 new decisions for the granting of one-time financial aid of 1,592 unemployed former employees and retrenched workers, of the following companies:- "Enomeni |
| ευρώ σε 1.592 ανέργους πρώην εργαζόμενους και εργαζόμενους σε επίσχεση, των εξής επιχειρήσεων: • «Ενωμένη Κλωστοϋφαντουργία Ανώνυμη Εταιρεία» σε 520 άτομα • «SHELMAN (ΣΕΛΜΑΝ) Α.Ε.» σε 391 άτομα • «Δημοσιογραφικός Οργανισμός Λαμπράκη (Δ.Ο.Λ.) Α.Ε.» και των θυγατρικών της: «ΔΟΛ DIGITAL ΑΝΩΝΥΜΗ ΕΤΑΙΡΕΙΑ» & «HEARST ΔΟΛ ΕΚΔΟΤΙΚΗ ΜΟΝΟΠΡΟΣΩΠΗ ΕΤΑΙΡΕΙΑ ΠΕΡΙΟΡΙΣΜΕΝΗΣ ΕΥΘΥΝΗΣ» σε 326 άτομα • «ΕΥΡΩΠΗ Πρακτορείο Διανομής Τύπου Α.Ε.» σε 101 άτομα • «ΧΟΝΤΟΣ ΠΑΛΛΑΣ ΠΟΛΥΚΑΤΑΣΤΗΜΑΤΑ ΑΝΩΝΥΜΗ ΕΤΑΙΡΕΙΑ» σε 91 άτομα • «ΕΞΠΡΕΣ – Δ.ΚΑΛΟΦΩΛΙΑΣ ΕΚΔΟΤΙΚΗ ΕΚΤΥΠΩΤΙΚΗ ΑΝΩΝΥΜΟΣ ΕΤΑΙΡΕΙΑ» σε 40 άτομα • «ΑΦΟΙ ΚΑΡΥΠΙΔΗ Α.Ε.» σε 40 άτομα • «Ραδιοφωνικές επιχειρήσεις ΒΗΜΑ FM Α.Ε.» σε 29 άτομα • «ΕΔΗΚΑ Α.Ε.» σε 25 άτομα • «ΕΥΒΟΙΑ MARKET ΠΡΟΜΗΘΕΥΤΙΚΗ» σε 13 άτομα • «METROPOLIS Α.Ε.Ε.», «ΠΑΝΔΩΡΑ Α.Ε.», «ΤΕΡΨΙΧΟΡΗ ΕΠΕ», «ΑΜΦΙΤΡΙΤΗ ΕΠΕ» (Όμιλος ΜΕΤΡΟΠΟΛΙΣ) σε 11 άτομα • «ΙΜΑΚΟ ΜΗΝΤΙΑ ΑΝΩΝΥΜΗ ΕΤΑΙΡΕΙΑ ΜΕΣΩΝ ΜΑΖΙΚΗΣ ΕΝΗΜΕΡΩΣΗΣ» σε 5 άτομα Το Υπουργείο Εργασίας, Κοινωνικής Ασφάλισης και Κοινωνικής Αλληλεγγύης, αξιοποιώντας κάθε διαθέσιμο εργαλείο, θα συνεχίσει να εργάζεται με γνώμονα την στήριξη της κοινωνίας και την επούλωση των πληγών της κρίσης. | Kloistofitourgia Anonimos Anonimos" 520 persons - "SHELMAN (SELMAN) S.A." 391 persons - 'Lambrakis Journalistic Organisation (D.O.L.) S.A.' and its subsidiaries: "DOL DIGITAL S.A." & "HEARST DOL EDITORIAL LIMITED LIABILITY COMPANY" persons - "EUROPE Press Distribution Agency S.A." 101 persons - 'HONTOS PALAS POLYKATISTS S.A.' 91 persons - 'EXPRES - D. KALOFOLIAS EDITORIAL PRINTING COMPANY S.A.' 40 persons - 'KARYPIDI BROTHERS S.A.' 40 persons - 'VIMA FM Radio Operations S.A.' 29 persons - 'EDIKA S.A.'in 25 persons - 'EVVOIA MARKET SUPPLY' in 13 persons - 'METROPOLIS S.A.', 'PANDORA S.A.' in 25 persons ", "TERPSICHORI LTD", "AMFITRITI LTD" (METROPOLIS Group) 11 persons - "IMAKO MINTIA S.A. Mass Media Company" 5 persons The Ministry of Labour, Social Security and Social Solidarity, using every available tool, will continue to work with a view to the support of society and the healing of the wounds of the crisis. |
| Human author summary | |
| "Συνολικά 12 νέες αποφάσεις για τη χορήγηση εφάπαξ οικονομικής ενίσχυσης, ύψους 1.000 ευρώ, σε 1.592 ανέργους πρώην εργαζομένους και εργαζομένους σε επίσχεση, υπέγραψε η Έφη Αχτσιόγλου." | "A total of 12 new decisions for the granting of one-off financial assistance of €1,000 to 1,592 unemployed former workers and retrenched workers were signed by Efi Achtsioglou." |
| Greek-mT5-small | |
| Το εργαλείο της εφάπαξ οικονομικής ενίσχυσης ύψους χιλίων ευρώ ενεργοποίησε το Υπουργείο Εργασίας, Κοινωνικής Ασφάλισης και Κοινωνικής Αλληλεγγύης. | The Ministry of Labour, Social Security and Social Solidarity has activated the tool of one-off financial support of one thousand euros. |
| Greek-umT5-small | |
| Η χορήγηση της εφάπαξ οικονομικής ενίσχυσης ύψους χιλίων ευρώ είχε ουσιαστικά ανασταλεί την περίοδο 2012-2014, καθώς η προηγούμενη κυβέρνηση ενέκρινε μόνο ένα σχετικό αίτημα σωματείου. | The granting of the one-off financial aid of one thousand euros was effectively suspended in 2012-2014, as the previous government approved only one relevant union request. |
| Greek-umT5-base | |
| Το υπουργείο Εργασίας ενεργοποίησε το εργαλείο της εφάπαξ οικονομικής ενίσχυσης ύψους χιλίων ευρώ, προς χιλιάδες ανέργους και εργαζόμενους σε επίσχεση. | The Ministry of Labour has activated the tool of one-off financial assistance of one thousand euros to thousands of unemployed and retrenched workers. |
| GreekBART | |
| Η χορήγηση της ενίσχυσης αυτής είχε ουσιαστικά ανασταλεί την περίοδο 2012-2014, καθώς η προηγούμενη κυβέρνηση ενέκρινε μόνο ένα σχετικό αίτημα σωματείου. | The granting of this aid was effectively suspended in 2012-2014, as the previous government approved only one relevant union request. |
| TextRank | |
| Α.Ε.» και των θυγατρικών της: «ΔΟΛ DIGI--TAL ΑΝΩΝΥΜΗ ΕΤΑΙΡΕΙΑ» & «HEARST ΔΟΛ ΕΚΔΟΤΙΚΗ ΜΟΝΟΠΡΟΣΩΠΗ ΕΤΑΙΡΕΙΑ ΠΕΡΙΟΡΙΣΜΕΝΗΣ ΕΥΘΥΝΗΣ» σε 326 άτομα • «ΕΥΡΩΠΗ Πρακτορείο Διανομής Α.Ε.» σε 101 άτομα • «ΧΟΝΤΟΣ ΠΑΛΛΑΣ ΠΟΛΥΚΑΤΑΣΤΗΜΑΤΑ ΑΝΩΝΥΜΗ ΕΤΑΙΡΕΙΑ» σε 91 άτομα • «ΕΞΠΡΕΣ – Δ.ΚΑΛΟΦΩΛΙΑΣ ΕΚΔΟΤΙΚΗ ΕΚΤΥΠΩΤΙΚΗ ΑΝΩΝΥΜΟΣ ΕΤΑΙΡΕΙΑ» σε 40 άτομα • «ΑΦΟΙ Ι ΚΑΡΥΠΙΔΗ Α.Ε.» σε 40 άτομα • «Ραδιοφωνικές επιχειρήσεις ΒΗΜΑ FM Α.Ε.» σε 29 άτομα • «ΕΔΗΚΑ Α.Ε.» σε 25 άτομα • «ΕΥΒΟΙΑ MARKET ΠΡΟΜΗΘΕΥΤΙΚΗ» σε 13 άτομα • «METROPOLIS Α.Ε.Ε.», «ΠΑΝΔΩΡΑ Α.Ε.», «ΤΕΡΨΙΧΟΡΗ ΕΠΕ», «ΑΜΦΙΤΡΙΤΗ ΕΠΕ» (Όμιλος ΜΕΤΡΟΠΟΛΙΣ) σε 11 άτομα • «ΙΜΑΚΟ ΜΗΝΤΙΑ ΑΝΩΝΥΜΗ ΕΤΑΙΡΕΙΑ ΜΕΣΩΝ ΜΑΖΙΚΗΣ ΕΝΗΜΕΡΩΣΗΣ» σε 5 άτομα Το Υπουργείο Εργασίας, Κοινωνικής Ασφάλισης και Κοινωνικής Αλληλεγγύης, αξιοποιώντας κάθε διαθέσιμο εργαλείο, θα συνεχίσει να εργάζεται με γνώμονα την στήριξη της κοινωνίας και την επούλωση των πληγών της κρίσης. | S.A.' and its subsidiaries: "DOL DIGITAL S.A." & "HEARST DOL EDITORIAL LIMITED LIABILITY COMPANY" persons - "EUROPE Press Distribution Agency S.A." 101 persons - 'HONTOS PALAS POLYKATISTS S.A.' 91 persons - 'EXPRES - D. KALOFOLIAS EDITORIAL PRINTING COMPANY S.A.' 40 persons - 'KARYPIDI BROTHERS S.A.' 40 persons - 'VIMA FM Radio Operations S.A.' 29 persons - 'EDIKA S.A.'in 25 persons - 'EVVOIA MARKET SUPPLY' in 13 persons - 'METROPOLIS S.A.', 'PANDORA S.A.' in 25 persons ", "TERPSICHORI LTD", "AMFITRITI LTD" (METROPOLIS Group) 11 persons - "IMAKO MINTIA S.A. Mass Media Company" 5 persons The Ministry of Labour, Social Security and Social Solidarity, using every available tool, will continue to work with a view to the support of society and the healing of the wounds of the crisis. |

**Example 9**

| Article | |
|---|---|
| Το Διεθνές Νομισματικό Ταμείο (ΔΝΤ) προβλέπει ένα χρέος ρεκόρ των πλούσιων χωρών το 2014 και κρίνει ""πιθανό"" να υπάρξει επιπλέον συμβολή των πιο εύπορων προσώπων και των πολυεθνικών επιχειρήσεων σε μια μείωση των ελλειμμάτων, σύμφωνα με έκθεσή του η οποία δόθηκε σήμερα στη δημοσιότητα. ""Φαίνεται ότι υπάρχει ένα επαρκές περιθώριο σε πολλές ανεπτυγμένες χώρες για να αντληθούν επιπλέον έσοδα από τα πιο υψηλά εισοδήματα"", υπογραμμίζει το ΔΝΤ στην έκθεσή του για την δημοσιονομική επιτήρηση. Κατά μέσον όρο, | The International Monetary Fund (IMF) predicts a record debt of rich countries in 2014 and considers it "likely" that the wealthiest individuals and multinational corporations will make an additional contribution to deficit reduction, according to a report released today. """It appears that there is sufficient scope in many developed countries to raise additional revenues from the highest incomes,""" the IMF underlines in its fiscal surveillance report. On average, developed countries' public debt is expected to reach an "all-time high" of 110% of their GDP in 2014, 35 |
| το δημόσιο χρέος των ανεπτυγμένων χωρών αναμένεται να φτάσει το ""ιστορικό υψηλό"" του 110% του ΑΕΠ τους το 2014, δηλαδή θα βρίσκεται 35 μονάδες πιο πάνω από το ποσοστό του 2007, επισημαίνει το ΔΝΤ στην έκθεσή του. Με μια αναλογία χρέους/ΑΕΠ της τάξης του 242,3% που προβλέπεται να έχει το 2014, η Ιαπωνία αναμένεται να βρίσκεται πρώτη στον κατάλογο των υπερχρεωμένων ανεπτυγμένων χωρών, ακολουθούμενη από την Ελλάδα (174%), την Ιταλία (133,1%) και την Πορτογαλία (125,3%). Οι ΗΠΑ, οι οποίες έχουν παραλύσει από ένα δημοσιονομικό αδιέξοδο και απειλούνται από μια πιθανή στάση πληρωμών, θα δουν το χρέος τους να ανεβαίνει στο 107,3% του ΑΕΠ τους το 2014, δηλαδή θα βρίσκονται πολύ πιο μπροστά από την Γαλλία και το 94,8% στο οποίο αναμένεται ότι θα ανέρχεται την ερχόμενη χρονιά το χρέος της. Η δεύτερη οικονομική δύναμη του κόσμου, η Κίνα δίνει την εικόνα του καλού μαθητή με μια αναλογία χρέους/ΑΕΠ μόνον 20,9% την ερχόμενη χρονιά, σύμφωνα με το ΔΝΤ. ""Παρά τις προόδους στη μείωση των ελλειμμάτων, οι δημοσιονομικές αδυναμίες παραμένουν βαθιές στις ανεπτυγμένες χώρες"", επισημαίνεται στην έκθεση. Απέναντι σε αυτές τις ανισορροπίες, το ΔΝΤ εκφράζει την ανησυχία του καθώς βλέπει ""ένα φορολογικό σύστημα υπό πίεση"", το οποίο ευνοεί τον ανταγωνισμό μεταξύ των κρατών και επιτρέπει στους εύπορους φορολογούμενους και στις πολυεθνικές να ελαφρύνουν τους φόρους τους. Μόνον στις ΗΠΑ, το ΔΝΤ υπολογίζει σε 60 δισεκατομμύρια δολάρια τα έσοδα που φέρεται ότι χάνονται λόγω τεχνικών βελτιστοποίησης της φορολογίας των πολυεθνικών. Το ΔΝΤ επισημαίνει ότι οι τελευταίες δεκαετίες έχουν σηματοδοτηθεί από μια ""θεαματική άνοδο"" του πλούτου του ""1%"" των πιο πλούσιων, κυρίως στον αγγλοσαξονικό κόσμο, χωρίς ωστόσο η φορολογία να έχει προσαρμοστεί σε αυτήν την εξέλιξη. ""Σε πολλές χώρες θα ήταν πιθανό να επιβληθούν επιπλέον φόροι σε αυτούς που διαθέτουν τα πιο υψηλά εισοδήματα"", υπογραμμίζει το ΔΝΤ, το οποίο κρίνει εξάλλου ""συνετό"" τον υπολογισμό σε 4.500 δισεκατομμύρια δολάρια των διαθεσίμων που αποκρύπτονται από ιδιώτες σε φορολογικούς παραδείσους. Οι χώρες της Ομάδας των Είκοσι (G20), οι υπουργοί Οικονομικών των οποίων συναντώνται αυτήν την εβδομάδα στην Ουάσινγκτον, ξεκίνησαν πρόσφατα πρωτοβουλίες για την πάταξη της φοροδιαφυγής. | points above the 2007 rate, the IMF said in its report. With a debt-to-GDP ratio of 242.3% projected for 2014, Japan is expected to top the list of indebted developed countries, followed by Greece (174%), Italy (133.1%) and Portugal (125.3%).The US, which are paralysed by a fiscal stalemate and threatened by a possible default, will see their debt rise to 107.3% of their GDP in 2014, well ahead of France and the 94.8% that its debt is expected to reach next year. The world's second economic powerhouse, China, gives the impression of being a good student with a debt/GDP ratio of only 20.9% next year, according to the IMF. "Despite advances in deficit reduction, fiscal weaknesses remain deep in developed countries," the report said. In the face of these imbalances, the IMF is concerned as it sees ""a tax system under pressure"", which favours competition between countries and allows wealthy taxpayers and multinationals to ease their taxes. In the US alone, the IMF estimates $60 billion in revenue allegedly lost due to tax optimisation techniques for multinationals. The IMF points out that the last decades have been marked by a "spectacular rise" in the wealth of the "1%" of the richest, especially in the Anglo-Saxon world, but that taxation has not adapted to this development. "In many countries it would be possible to impose additional taxes on those with the highest incomes," stresses the IMF, which also considers it 'prudent' to estimate at 4.500 billion dollars the assets hidden by individuals in tax havens. The G20 countries, whose finance ministers are meeting this week in Washington, have recently launched initiatives to combat tax evasion. |
| Human author summary | |
| Επιπλέον συμβολή των εύπορων προσώπων και των πολυεθνικών επιχειρήσεων σε μείωση των ελλειμμάτων βλέπει το ΔΝΤ, το οποίο προβλέπει και χρέος ρεκόρ των πλούσιων χωρών. | An additional contribution of wealthy individuals and multinational corporations to deficit reduction is seen by the IMF, which also predicts record debt for rich countries. |
| Greek-mT5-small | |
| Το ΔΝΤ προβλέπει ένα χρέος ρεκόρ των πλούσιων χωρών το 2014. Τι αναφέρει για την πορεία των ελλειμμάτων. | The IMF predicts a record debt for rich countries in 2014. What does it say about the path of deficits. |
| Greek-umT5-small | |
| Το ΔΝΤ προβλέπει χρέος ρεκόρ των πλούσιων χωρών το 2014 και κρίνει "πιθανό" να υπάρξει συμβολή των πιο εύπορων προσώπων και των πολυεθνικών επιχειρήσεων σε μια μείωση των ελλειμμάτων. | The IMF predicts a record debt of rich countries in 2014 and considers it "likely" that the wealthiest individuals and multinational corporations will contribute to a reduction in deficits. |
| Greek-umT5-base | |
| Χρέος ρεκόρ για τις πλούσιες χώρες το 2014 προβλέπει το ΔΝΤ. Πιθανό να υπάρξει επιπλέον συμβολή των πιο εύπορων προσώπων και των πολυεθνικών επιχειρήσεων σε μια μείωση των ελλειμμάτων. | Record debt for rich countries in 2014, the IMF predicts. There is likely to be an additional contribution from the wealthiest individuals and multinational corporations to a reduction in deficits. |
| GreekBART | |
| Το Διεθνές Νομισματικό Ταμείο (ΔΝΤ) προβλέπει ένα χρέος ρεκόρ των πλούσιων χωρών το 2014. | The International Monetary Fund (IMF) predicts a record level of rich country debt in 2014. |
| TextRank | |
| Κατά μέσον όρο, το δημόσιο χρέος των ανεπτυγμένων χωρών αναμένεται να φτάσει το ""ιστορικό υψηλό"" του 110% του ΑΕΠ τους το 2014, δηλαδή θα βρίσκεται 35 μονάδες πιο πάνω από το ποσοστό του 2007, επισημαίνει το ΔΝΤ στην έκθεσή του. | On average, developed countries' public debt is expected to reach an "all-time high" of 110% of their GDP in 2014, 35 points above the 2007 rate, the IMF said in its report. |